\documentclass[sigconf]{acmart}

\AtBeginDocument{%
  \providecommand\BibTeX{{%
    \normalfont B\kern-0.5em{\scshape i\kern-0.25em b}\kern-0.8em\TeX}}}

\renewcommand\footnotetextcopyrightpermission[1]{}

\acmConference[Everything Is a Robot 
(and Nothing Is), CHI '26]{ACM Conference}{April 15, 2026}{Barcelona, Spain}%

\begin{document}

\title{Architectural HRI: Towards a Robotic Paradigm Shift in Human-Building Interaction}

\author{Alex Binh Vinh Duc Nguyen}
\email{alexbvd.nguyen@uantwerpen.be}
\orcid{0000-0001-5026-474X}
\affiliation{
  \institution{University of Antwerp}
  \city{Antwerp}
  \country{Belgium}
}
\affiliation{
  \institution{KU Leuven}
  \city{Leuven}
  \country{Belgium}
}


\begin{abstract}
Recent advances in sensing, communication, interfaces, control, and robotics are expanding Human-Building Interaction (HBI) beyond adaptive building services and facades toward the physical actuation of architectural space. In parallel, research in robotic furniture, swarm robotics, and shape-changing spaces shows that architectural elements can now be robotically augmented to move, reconfigure, and adapt space. We propose that these advances promise a paradigm shift in HBI, in which multiple building layers physically adapt in synchrony to support occupant needs and sustainability goals more holistically. Conversely, we argue that this emerging paradigm also provides an ideal case for transferring HRI knowledge to unconventional robotic morphologies, including the interpretation of the robot as multiple architectural layers or even as a building. However, this research agenda remains challenged by the temporal, spatial, and social complexity of architectural HRI, and by fragmented knowledge across HCI, environmental psychology, cognitive science, and architecture. We therefore call for interdisciplinary research that unifies the why, what, and how of robotic actuation in architectural forms.
\end{abstract}


\begin{teaserfigure}
  \centering
  \includegraphics[width=0.85\textwidth]{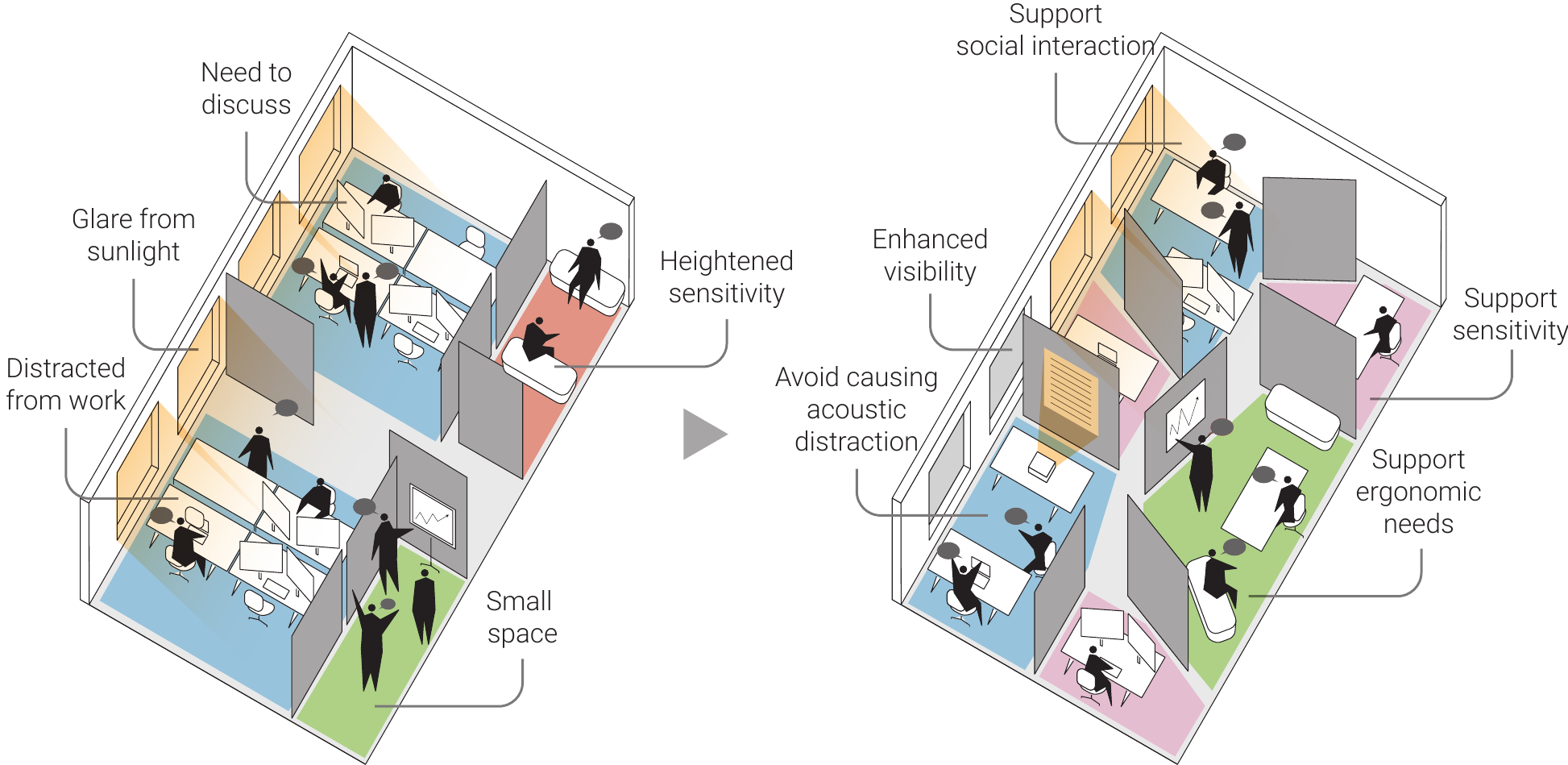}
  \caption{The robotic paradigm shift in HBI envisions multiple layers of a building working in synchrony to support the needs of co-located occupants. In this example shared office, these layers include mobile robotic partitions (\textit{"space plan"}) and desks (\textit{"stuffs"}), smart window shading systems (\textit{"skin"}), and smart lighting (\textit{"services"}), all coordinating their adaptations based on the individual needs and preferences of multiple occupants.}
  \label{fig:orchestration}
\end{teaserfigure}

\maketitle

\section{Human-Building Interaction (HBI)}

Recent technological advances have enabled a new generation of smart building systems. Enhanced sensing, communication, interfaces, and control technologies now allow for new forms of interaction between buildings and their occupants~\cite{Bier2014,Fox2016,Green2016}. The field of Human-Building Interaction (HBI) investigates these advances, informing how buildings are conceived, designed, constructed, inhabited, and managed to improve the lives and experiences of their occupants~\cite{Alavi2019,BecerikGerber2022,Wiberg2020, Rao2025}. Historically, HBI research has focused on advancing automated building management technologies, with an emphasis on reducing energy consumption while maintaining thermal comfort~\cite{BecerikGerber2022}. This focus has led to the development of 'adaptive' facades~\cite{Navarro2020,Tabadkani2021} and smart building service systems~\cite{Aliero2022,Vijayan2020} that can adjust Indoor Environmental Quality (IEQ) based on the activities, personalities, or preferences of occupants.

Despite these advances, recent post-occupancy evaluations have consistently reported that many occupants remain dissatisfied with their living~\cite{Chen2023} or working~\cite{Parkinson2023} spaces. This persistent dissatisfaction stems from the fact that occupants' spatial needs are highly situational, shifting between activities~\cite{Lindberg2018} and differing significantly between individuals sharing the same space~\cite{Hong2020}. Yet, beyond the \textit{service} and \textit{facade} layers, the other shearing layers of the built environment~\cite{Brand1995}, ranging from \textit{structure} to \textit{layout} and \textit{furniture}, remain largely static and are unable to support the dynamic and situational needs of occupants.

\section{Human-Robot Interaction (HRI)}

From the domain of architectural design, the field of adaptive architecture investigates how digital and robotic technology can automatically and autonomously adapt an architectural space to meet the changing needs of occupants. During five decades of research, visionaries in this field have proposed various theoretical manifestos~\cite{Fox2016} and provocative installations~\cite{Oosterhuis2008}, which demonstrate the potential of adaptive layouts not only to host multiple functional purposes, but also to evoke compelling architectural experiences~\cite{Achten2013, Meagher2015}. Thanks to recent technological advancements, this vision of adaptive layout has become much closer to reality. Evidently, a growing number of innovative start-ups like Ori\footnote{Ori: \href{https://www.oriliving.com/}{oriliving.com}}, Beyome\footnote{Beyome: \href{https://www.beyome.live/language/en/}{beyome.live}}, or Bumblebee\footnote{Bumblebee: \href{https://bumblebeespaces.com/}{bumblebeespaces.com}} are now offering robotic solutions to optimise space use within apartments by flexibly moving large furniture items like beds or sofas between their usual locations and storage spaces nestled in underutilised areas, such as beneath the floor, above ceilings, or inside space-delimiting cupboards.

In parallel, advances in Human-Robot Interaction (HRI), particularly in robotic furniture~\cite{Ju2009, Lello2011, Sirkin2015}, hold the potential to initiate a new form of HBI through the physical actuation of the built environment at an architectural scale~\cite{Bier2014,Green2016}. Research in swarm robotics~\cite{Onishi2022}, robotic furniture~\cite{Sirkin2015}, and shape-changing spaces~\cite{Jens2017} has demonstrated that everyday architectural elements, such as chairs~\cite{Agnihotri2019}, sofas~\cite{Spadafora2016}, benches~\cite{Yu2026}, tables~\cite{Takashima2015}, or walls~\cite{Nguyen2021} can now be robotically augmented to purposefully move, change shape, or re-arrange into different spatial configurations, or even gesture understandable intents toward occupants to nudge their behaviours.

\section{Architectural HRI: Towards a Robotic Paradigm Shift in HBI}

\begin{figure}[ht]
    \centering
    \includegraphics[width=\linewidth]{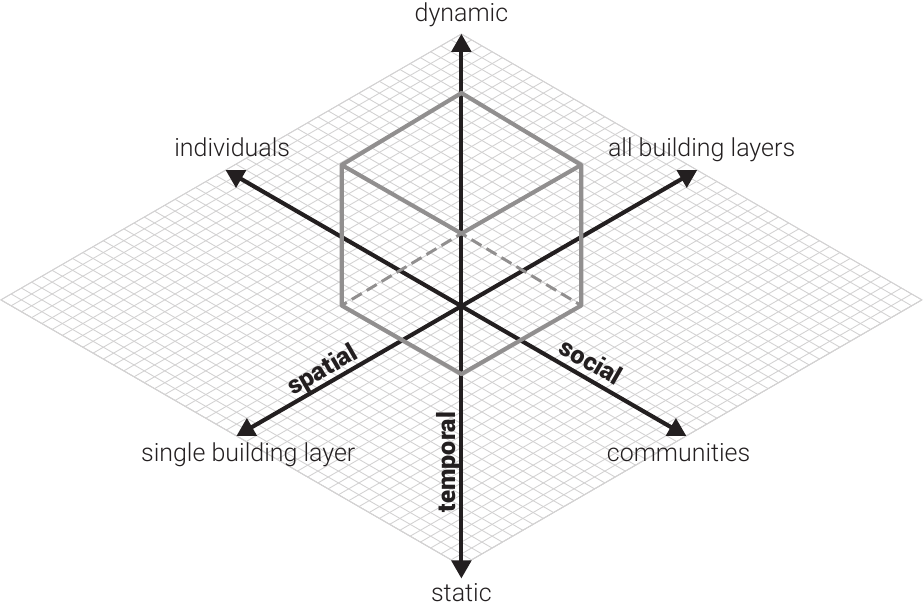}
    \caption{"Architectural HBI" must operate across three dimensions: \textit{temporal} (from static space to dynamic actuation), \textit{spatial} (from a single building layer to the whole building), and \textit{social} (from individual occupants to whole building communities).}
    \label{fig:dimension}
    \vspace{-15pt}
\end{figure}

We propose that recent robotic advances promise a paradigm shift in HBI, in which multiple layers of the built environment can physically adapt in synchrony to support occupant needs, as exemplified in \autoref{fig:orchestration}. This robotic paradigm shift holds the potential for HBI to support both occupant needs and sustainability goals in more holistic ways. Physically reconfiguring spatial layouts and shapes can not only alter spatial qualities such as visibility and accessibility~\cite{Nguyen2024}, but also potentially influence overall IEQ by directing airflow or manipulating spatial volumes. Through robotic actuation, HBI can also operate at a much finer-grained level, for example by individually optimising sub-spaces tailored to the needs of co-located occupants~\cite{Balci2025}, or by dynamically reconfiguring layouts to accommodate changing activities throughout the day~\cite{Nguyen2022}.

Conversely, we argue that the emerging robotic paradigm in HBI represents an ideal case for the transfer of HRI knowledge to unconventional robotic morphologies, i.e. "\textit{robot as multiple architectural layers}", or even "\textit{robot as a building}". Our own research shows that this lens provides an alternative interpretation of HRI, shifting the focus from the interaction of humans with robotic motion to also considering its spatial impact~\cite{Nguyen2025, Balci2025, Yu2026}. This interlink provides a bridge between HRI and HBI, potentially motivating a more seamless integration of robots into everyday environments.

Nonetheless, this research agenda still faces several open challenges, particularly due to the multifaceted nature of "\textit{architectural HRI}", i.e. the interaction of humans with robots in architectural morphologies, which must operate across \textit{temporal}, \textit{spatial}, and \textit{social} dimensions, as shown in \autoref{fig:dimension}. For instance, architectural HRIs can unfold both longitudinally and instantaneously, as the built environment can alternate between serving as a static backdrop for everyday activities and functioning as an adaptive system that responds instantly to occupant needs~\cite{Nguyen2023}. Architectural HRIs may also vary widely in spatial scale, from a single building layer, to multiple building layers, or even an entire building, each carrying distinct social implications for individuals, groups, and communities~\cite{Nguyen2024Habitech}.

Although recent research has provided early insights into how such architectural HRIs should be designed, these insights remain largely fragmented across disciplines such as Human-Computer Interaction (HCI), environmental psychology, cognitive science, and architecture, among others. Because this paradigm shift requires the unification of meaningful robotic actuation in physical architectural forms (the "\textbf{why}") via software or interfaces (the "\textbf{what}") to support social processes and human interactions (the "\textbf{how}"), we call for interdisciplinary research efforts to account for the complex interplay of technological, cognitive, and environmental factors~\cite{BecerikGerber2022,VandeMoere2024}.


\bibliographystyle{ACM-Reference-Format}
\balance
\bibliography{sample-base}

@book{Fox2016,
  title={Interactive Architecture: Adaptive World},
  author={Fox, Michael},
  year={2016},
  publisher={Chronicle Books},
  isbn={9781616895112},
  address={San Francisco, California, United States}
}

@inproceedings{Achten2013,
author = {Achten, Henri},
booktitle = {Computation and Performance – Proceedings of the 31st eCAADe Conference},
pages = {477--485},
title = {{Buildings with an Attitude: Personality traits for the design of interactive architecture}},
volume = {1},
year = {2013},
doi={10.52842/conf.ecaade.2013.1.477},
publisher={eCAADe},
address={Delft, Netherlands}
}

@article{Meagher2015,
author = {Meagher, Mark},
doi = {10.1016/j.foar.2015.03.002},
issn = {2095-2635},
journal = {Frontiers of Architectural Research},
number = {2},
pages = {159-165},
title = {{Designing for Change: The Poetic Potential of Responsive Architecture}},
volume = {4},
year = {2015}
}

@book{Green2016,
  title={Architectural Robotics: Ecosystems of Bits, Bytes, and Biology},
  author={Keith Evan Green},
  isbn={9780262033954},
  lccn={2015037895},
  series={The MIT Press},
  year={2016},
  publisher={MIT Press},
  address = {Cambridge, MA, USA}
}

@inproceedings{Spadafora2016,
author = {Spadafora, Marco and Chahuneau, Victor and Martelaro, Nikolas and Sirkin, David and Ju, Wendy},
title = {Designing the Behavior of Interactive Objects},
year = {2016},
isbn = {9781450335829},
publisher = {ACM},
address = {New York, NY, USA},
doi = {10.1145/2839462.2839502},
booktitle = {Proceedings of the TEI '16: Tenth International Conference on Tangible, Embedded, and Embodied Interaction},
pages = {70–77},
numpages = {8},
location = {Eindhoven, Netherlands},
series = {TEI '16}
}

@inproceedings{Agnihotri2019,
author = {Agnihotri, Abhijeet and Knight, Heather},
title = {Persuasive ChairBots: A (Mostly) Robot-Recruited Experiment},
year = {2019},
publisher = {IEEE Press},
doi = {10.1109/RO-MAN46459.2019.8956262},
booktitle = {2019 28th IEEE International Conference on Robot and Human Interactive Communication (RO-MAN)},
pages = {1–7},
numpages = {7},
address = {New Delhi, India}
}

@inproceedings{Nguyen2024,
author = {Nguyen, Binh Vinh Duc and Vande Moere, Andrew},
title = {The Adaptive Architectural Layout: How the Control of a Semi-Autonomous Mobile Robotic Partition was Shared to Mediate the Environmental Demands and Resources of an Open-Plan Office},
year = {2024},
publisher = {ACM},
address = {New York, NY, USA},
doi = {10.1145/3613904.3642465},
ISBN = {979-8-4007-0330-0/24/05},
booktitle = {Proceedings of the 2024 CHI Conference on Human Factors in Computing Systems},
location = {Honolulu, HI, USA},
series = {CHI '24},
numpages = {20},
}

@inproceedings{Nguyen2021,
author = {Binh Vinh Duc Nguyen and Adalberto L. Simeone and Andrew Vande Moere},
title = {Exploring an Architectural Framework for Human-Building Interaction via a Semi-Immersive Cross-Reality Methodology},
year = {2021},
publisher = {ACM},
doi = {10.1145/3434073.3444643},
booktitle = {Proceedings of the 2021 ACM/IEEE International Conference on Human-Robot Interaction},
numpages = {10},
location = {Boulder, CO, USA},
address = {New York, NY, USA},
series = {HRI '21}
}

@inproceedings{Sirkin2015,
author = {Sirkin, David and Mok, Brian and Yang, Stephen and Ju, Wendy},
title = {Mechanical Ottoman: How Robotic Furniture Offers and Withdraws Support},
year = {2015},
isbn = {9781450328838},
publisher = {ACM},
address = {New York, NY, USA},
doi = {10.1145/2696454.2696461},
booktitle = {Proceedings of the Tenth Annual ACM/IEEE International Conference on Human-Robot Interaction},
pages = {11–18},
numpages = {8},
location = {Portland, Oregon, USA},
series = {HRI '15}
}

@inproceedings{Onishi2022,
author = {Onishi, Yuki and Takashima, Kazuki and Higashiyama, Shoi and Fujita, Kazuyuki and Kitamura, Yoshifumi},
title = {WaddleWalls: Room-Scale Interactive Partitioning System Using a Swarm of Robotic Partitions},
year = {2022},
isbn = {9781450393201},
publisher = {ACM},
address = {New York, NY, USA},
url = {https://doi.org/10.1145/3526113.3545615},
doi = {10.1145/3526113.3545615},
booktitle = {Proceedings of the 35th Annual ACM Symposium on User Interface Software and Technology},
articleno = {29},
numpages = {15},
location = {Bend, OR, USA},
series = {UIST '22}
}

@article{Wiberg2020,
author = {Wiberg, Mikael},
title = {Interaction and Architecture is Dead.: Long Live Architectural Interactivity!},
year = {2020},
issue_date = {March - April 2020},
publisher = {ACM},
address = {New York, NY, USA},
volume = {27},
number = {2},
issn = {1072-5520},
doi = {10.1145/3378567},
journal = {Interactions},
month = feb,
pages = {72–75},
numpages = {4}
}

@article{Navarro2020,
title = {Occupant-Facade interaction: a review and classification scheme},
journal = {Building and Environment},
volume = {177},
pages = {106880},
year = {2020},
issn = {0360-1323},
doi = {https://doi.org/10.1016/j.buildenv.2020.106880},
author = {Alessandra Luna-Navarro and Roel Loonen and Miren Juaristi and Aurora Monge-Barrio and Shady Attia and Mauro Overend},
}

@article{BecerikGerber2022,
title = {Ten Questions Concerning Human-Building Interaction Research for Improving the Quality of Life},
journal = {Building and Environment},
volume = {226},
pages = {109681},
year = {2022},
issn = {0360-1323},
doi = {10.1016/j.buildenv.2022.109681},
author = {Burçin Becerik-Gerber and Gale Lucas and Ashrant Aryal and Mohamad Awada and Mario Bergés and Sarah L Billington and Olga Boric-Lubecke and Ali Ghahramani and Arsalan Heydarian and Farrokh Jazizadeh and Ruying Liu and Runhe Zhu and Frederick Marks and Shawn Roll and Mirmahdi Seyedrezaei and John E. Taylor and Christoph Höelscher and Azam Khan and Jared Langevin and Matthew Louis Mauriello and Elizabeth Murnane and Haeyoung Noh and Marco Pritoni and Davide Schaumann and Jie Zhao},
}

@article{Parkinson2023,
  title={Common Sources of Occupant Dissatisfaction with Workspace Environments in 600 Office Buildings},
  author={Parkinson, Thomas and Schiavon, Stefano and Kim, Jungsoo and Betti, Giovanni},
  journal={Buildings and Cities},
  volume={4},
  number={1},
  year={2023},
  publisher={Ubiquity Press},
  pages={17-35},
}

@article {Lindberg2018,
	author = {Lindberg, Casey M and Srinivasan, Karthik and Gilligan, Brian and Razjouyan, Javad and Lee, Hyoki and Najafi, Bijan and Canada, Kelli J and Mehl, Matthias R and Currim, Faiz and Ram, Sudha and Lunden, Melissa M and Heerwagen, Judith H and Kampschroer, Kevin and Sternberg, Esther M},
	title = {Effects of office workstation type on physical activity and stress},
	volume = {75},
	number = {10},
	pages = {689--695},
	year = {2018},
	doi = {10.1136/oemed-2018-105077},
	publisher = {BMJ Publishing Group Ltd},
	issn = {1351-0711},
	journal = {Occupational and Environmental Medicine}
}

@article{Nguyen2022,
author = {Nguyen, Binh Vinh Duc and Han, Jihae and Vande Moere, Andrew},
title = {Towards Responsive Architecture That Mediates Place: Recommendations on How and When an Autonomously Moving Robotic Wall Should Adapt a Spatial Layout},
year = {2022},
issue_date = {November 2022},
publisher = {Association for Computing Machinery},
address = {New York, NY, USA},
volume = {6},
number = {CSCW2},
doi = {10.1145/3555568},
journal = {Proc. ACM Hum.-Comput. Interact.},
month = {nov},
articleno = {467},
numpages = {27}
}

@article{Oosterhuis2008,
author = {Oosterhuis, Kas and Biloria, Nimish},
title = {Interactions with Proactive Architectural Spaces: The Muscle Projects},
year = {2008},
issue_date = {June 2008},
publisher = {ACM},
address = {New York, NY, USA},
volume = {51},
number = {6},
issn = {0001-0782},
doi = {10.1145/1349026.1349041},
journal = {Communications of the ACM},
month = jun,
pages = {70–78},
numpages = {9}
}

@article{Hong2020,
title = {Linking Human-Building Interactions in Shared Offices With Personality Traits},
journal = {Building and Environment},
volume = {170},
pages = {106602},
year = {2020},
issn = {0360-1323},
doi = {10.1016/j.buildenv.2019.106602},
author = {Tianzhen Hong and Chien-fei Chen and Zhe Wang and Xiaojing Xu},
}

@article{Tabadkani2021,
title = {A review of occupant-centric control strategies for adaptive facades},
journal = {Automation in Construction},
volume = {122},
pages = {103464},
year = {2021},
issn = {0926-5805},
doi = {10.1016/j.autcon.2020.103464},
author = {Amir Tabadkani and Astrid Roetzel and Hong Xian Li and Aris Tsangrassoulis},
}

@article{Bier2014,
 title={Robotic Building(s)},
 author={Henriette H. Bier},
 journal={Next Generation Building},
 year={2014},
 volume={1},
 number={1},
 doi={10.7564/14-NGBJ8},
 pages={10}
}

@inproceedings{Jens2017,
author = {Gr\o{}nb\ae{}k, Jens Emil and Korsgaard, Henrik and Petersen, Marianne Graves and Birk, Morten Henriksen and Krogh, Peter Gall},
title = {Proxemic Transitions: Designing Shape-Changing Furniture for Informal Meetings},
year = {2017},
isbn = {9781450346559},
publisher = {ACM},
address = {New York, NY, USA},
url = {https://doi.org/10.1145/3025453.3025487},
booktitle = {Proceedings of the 2017 CHI Conference on Human Factors in Computing Systems},
pages = {7029–7041},
numpages = {13},
series = {CHI '17}
}

@article{Lello2011,
author = {Enrico Di Lello and Alessandro Saffiotti},
title = {The PEIS Table: An Autonomous Robotic Table for Domestic Environments},
journal = {Automatika},
volume = {52},
number = {3},
pages = {244-255},
year  = {2011},
publisher = {Taylor \& Francis},
doi = {10.1080/00051144.2011.11828423},
}

@article{Ju2009,
author = {Wendy Ju and Leila Takayama},
title = {Approachability: How People Interpret Automatic Door Movement as Gesture},
year = {2009},
publisher = {IJDesign},
address = {Taipei, Taiwan},
volume = {3},
number = {2},
url = {http://www.ijdesign.org/index.php/IJDesign/article/view/574/262},
journal = {International Journal of Design},
pages = {77--86}
}

@InProceedings{Takashima2015,
author="Takashima, Kazuki
and Asari, Yusuke
and Yokoyama, Hitomi
and Sharlin, Ehud
and Kitamura, Yoshifumi",
editor="Abascal, Julio
and Barbosa, Simone
and Fetter, Mirko
and Gross, Tom
and Palanque, Philippe
and Winckler, Marco",
title="MovemenTable: The Design of Moving Interactive Tabletops",
booktitle="Human-Computer Interaction -- INTERACT 2015",
year="2015",
publisher="Springer International Publishing",
address="Cham, Switzerland",
pages="296--314",
isbn="978-3-319-22698-9",
doi={10.1007/978-3-319-22698-9_19}
}


\end{document}